\documentclass[letterpaper, 10 pt, conference]{ieeeconf}  % Comment this line out if you need a4paper

\IEEEoverridecommandlockouts                              % This command is only needed if 
\usepackage{amsmath}
\usepackage{algorithm}
\usepackage{color, graphicx}% Alan: begin the font trial
\usepackage{booktabs, multirow}
\usepackage{tabularx}
\usepackage[table]{xcolor}
\usepackage{caption}
\usepackage{balance}
\usepackage[justification=raggedright]{caption}

\title{\LARGE \bf
YOLOStereo3D: A Step Back to 2D for Efficient Stereo 3D Detection
}

\author{Yuxuan Liu$^{1}$, Lujia Wang$^{2}$ and Ming Liu$^{1}$% <-this % stops a space
\thanks{*This work was supported by the National Natural Science Foundation
of China (Grant No. U1713211), the Research Grant Council of Hong
Kong SAR Government, China, under Project No. 11210017, and No.
21202816, and Shenzhen Science, Technology and Innovation Comission
(SZSTI) JCYJ20160428154842603, awarded to Prof. Ming Liu. And it was supported
by the Guangdong Science and Technology Plan Guangdong-Hong Kong
Cooperation Innovation Platform (Grant Number 2018B050502009) awarded
to Lujia Wang. (Lujia Wang is the corresponding author).} 
\thanks{$^{1}$Yuxuan Liu and Ming Liu are with the Robotics and Multi-Perception Laborotary, 
Department of Electronic and Computer Engineering, The Hong Kong University of Science and Technology
        {\tt\footnotesize yliuhb@connect.ust.hk ,eelium@ust.hk}}%
\thanks{$^{2} $Lujia Wang is with Cloud Computing Lab of Shenzhen Institutes
of Advanced Technology, Chinese Academy of Sciences, China.
        {\tt\footnotesize lj.wang1@siat.ac.cn}}%
}

\begin{document}

\maketitle
\thispagestyle{empty}
\pagestyle{empty}

%%%%%%%%%%%%%%%%%%%%%%%%%%%%%%%%%%%%%%%%%%%%%%%%%%%%%%%%%%%%%%%%%%%%%%%%%%%%%%%%
\begin{abstract}

        Object detection in 3D with stereo cameras is an important problem in computer vision, and is particularly crucial in low-cost autonomous mobile robots without LiDARs. 
        Nowadays, most of the best-performing frameworks for stereo 3D object detection are based on dense depth reconstruction from disparity estimation, making them extremely computationally expensive.
        To enable real-world deployments of vision detection with binocular images, we take a step back to gain insights from 2D image-based detection frameworks and enhance them with stereo features.
        We incorporate knowledge and the inference structure from real-time one-stage 2D/3D object detector and introduce a light-weight stereo matching module. 
        Our proposed framework, YOLOStereo3D, is trained on one single GPU and runs at more than ten fps. It demonstrates performance comparable to state-of-the-art stereo 3D detection frameworks without usage of LiDAR data. The code will be published in https://github.com/Owen-Liuyuxuan/visualDet3D.

\end{abstract}

%%%%%%%%%%%%%%%%%%%%%%%%%%%%%%%%%%%%%%%%%%%%%%%%%%%%%%%%%%%%%%%%%%%%%%%%%%%%%%%%

%                                   % the textheight of the last page by a suitable amount.
%                                   % This command does not take effect until the next page
%                                   % so it should come on the page before the last. Make
%                                   % sure that you do not shorten the textheight too much.

%%%%%%%%%%%%%%%%%%%%%%%%%%%%%%%%%%%%%%%%%%%%%%%%%%%%%%%%%%%%%%%%%%%%%%%%%%%%%%%%

\section{Introduction}
\label{section:Introduction}
3D object detection is a fundamental problem in computer vision, and a crucial engineering problem for autonomous vehicles and mobile robots \cite{yang18HDNet_CORL} \cite{Zhou2019CORL}. 
With two horizontally-aligned RGB cameras with known displacement, it is possible to estimate depth by triangulation according to the pin-hole camera model.
Though this is a non-direct measurement for depth and is, in most cases, less robust than LiDAR-based approaches, the binocular setup is generally much cheaper and it is promising for low-cost applications such as mobile robots and autonomous logistic vehicles.
\begin{figure*}
    \centering
    \includegraphics[{width=0.95\textwidth}]{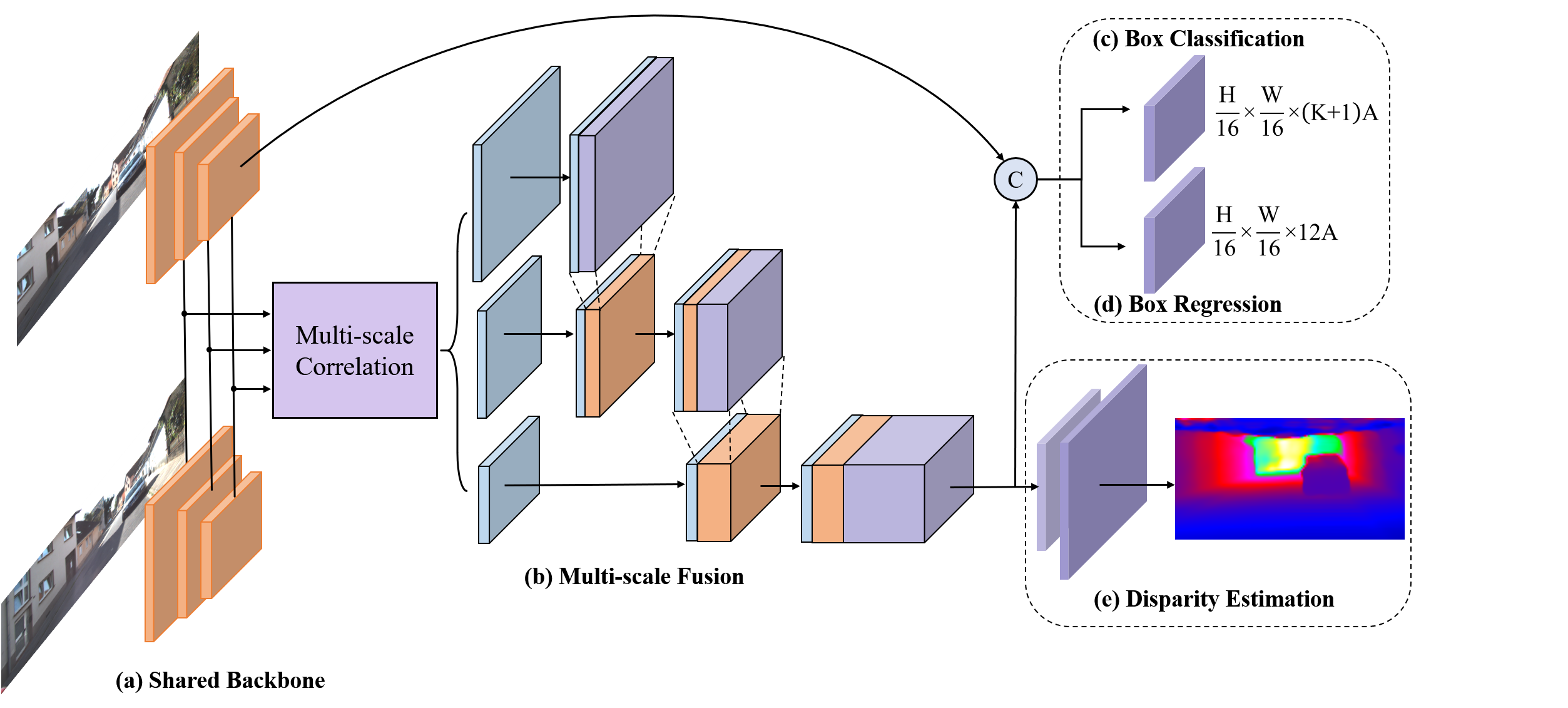}
        
    \caption{Network inference structure of YOLOStereo3D.
    YOLOStereo3D extracts multi-scale features from binocular images with a backbone network (a). 
    These features are passed through a multi-scale stereo matching and fusion module (b) as described in Section~\ref{subsec: multiscale_fusion}.
    Finally, the fused features are concatenated with the last feature from the left image and sent to the classification/regression branch to densely predict the 3D bounding boxes (c/d).
    The network also produces a disparity estimation during training (e).
   }
    \label{fig:architecture}
\end{figure*}

Many of the state-of-the-art frameworks in stereo 3D object detection stem from the idea of pseudo-LiDAR and are motivated by general stereo matching algorithms.
However, in 3D object detection, the model should focus on foreground objects. It is expected to be as accurate as possible since a disparity error of one or two pixels would cause a large error in terms of real-world distance.
Many researchers have delved deep into these problems to improve the performance of pseudo-LiDAR-based algorithms; some directly fine-tune the estimation of point clouds to improve performance \cite{You2019PLPP},\cite{Qian2020PLE2E}, while others utilize instance segmentation to focus the stereo matching network on foreground pixels \cite{Sun2020DispRCNN}, \cite{xu2020Zoomnet}, \cite{Pon2019OCStereo}.
However, a high-performance disparity estimation network, e.g., PSMNet \cite{Chang2018PSMNet}, usually takes more than 300 ms per frame on modern hardware on the KITTI dataset \cite{Geiger2012KITTI} and requires a huge GPU memory to train.
These issues hinder the deployment of stereo systems on low-cost robotic applications.

Many of the works mentioned above have shown in practice that transforming images into 3D features is usually sub-optimal and computationally expensive.
To improve the efficiency of stereo 3D detection algorithms while maintaining as much of their performance as possible, we propose selecting a different architecture.
Instead of casting the problem as a 3D detection problem with less accurate point clouds, we take a step back and treat it as a monocular 3D detection task with enhanced stereo features, which is the fundamental motivation of this work. 

The framework we propose, YOLOStereo3D, is a light-weight one-stage stereo 3D detection network (Section~\ref{sec.mono_anchor}).
To efficiently produce powerful stereo features, we re-introduce the pixel-wise correlation module to construct the cost-volume, instead of the popular concatenation-based module (Section~\ref{sec.costvolume}). 
Such a module produces a thin 2D feature map where each channel corresponds to a disparity hypothesis in stereo matching.
We then apply this module hierarchically to efficiently produce stereo features as 2D feature maps (Section~\ref{sec.ghost}), and we densely fuse these features (Section~\ref{sec.hierachical}) to form the base-feature of detection heads. The network is trained end-to-end without the use of LiDAR data (Section~\ref{sec.training}).

The main contributions of this paper are three-fold. 
\begin{itemize}
    \item For the inference architecture, we incorporate and optimize the inference pipeline from one-stage monocular 3D detection into stereo 3D detection.
    \item For the design of the network, we introduce a point-wise correlation module in stereo detection tasks and propose a hierarchical, densely-connected structure to utilize stereo features from multiple scales.
    \item For the experimental results, the proposed YOLOStereo3D produces competitive results on the KITTI 3D benchmark without using point clouds and with an inference time of less than 0.1 seconds per frame.
\end{itemize}

\section{Related Works}
\label{section:Relate}
\subsection{Stereo Matching}

Stereo matching algorithms focus on estimating the disparity between binocular images.
The current state-of-the-art frameworks for stereo matching apply siamese networks for feature extraction from two images, and construct 3D cost volumes to search the disparity value on each pixel exhaustively.
Early research applied the dot-product between binocular feature maps, with the resulting correlation directly forming an estimation of the disparity distribution \cite{Luo2016EffiStereo} \cite{ZbontarL15DotStereo}.
PSMNet \cite{Chang2018PSMNet} and GCNet \cite{Kendall2017GCNet} constructed concatenation-based cost volumes and applied multiple 3D convolutions to produce disparity outputs.
The recent FADNet managed to perform a fast stereo estimation with a point-wise correlation module \cite{wang2020FADNet}.
Zhang \textit{et al.} proposed stereo focal loss to improve the loss function formulation in disparity estimation \cite{Zhang2019AcfNet}.
Our work, similar to many other stereo 3D object detection algorithms, is developed upon these studies and utilizes the stereo matching features to boost detection performance.

\subsection{Visual 3D Object Detection}

\subsubsection{Stereo 3D Object Detection}

Stereo 3D object detection is usually considered as a tractable but computationally hard problem.
Recent advances in stereo 3D object detection algorithms are based on the idea of pseudo-LiDAR \cite{wang2018pseudo}.
DispRCNN \cite{Sun2020DispRCNN}, ZoomNet \cite{xu2020Zoomnet}, and OC Stereo \cite{Pon2019OCStereo} applied instance segmentation on binocular images to construct a local point cloud for each detected instance to improve the accuracy of disparity estimation on foreground objects.
Pseudo-LiDAR++ \cite{You2019PLPP} recognized that uniform 3D convolution might not be suitable to process the disparity cost volume, and transformation to the depth cost volume may be needed.

We point out that all the aforementioned algorithms require more than 0.3 seconds runtime per frame. Moreover,
Pseudo-LiDAR++ \cite{You2019PLPP}, ZoomNet \cite{xu2020Zoomnet}, OC Stereo \cite{Pon2019OCStereo} and DSGN \cite{Chen2020DSGN} required point cloud data during training or need point cloud data to help the training process.
DispRCNN \cite{Sun2020DispRCNN} and the baseline Pseudo-LiDAR \cite{wang2018pseudo} required off-the-shelf disparity modules, which are usually trained with depth images or point cloud data.

YOLOStereo3D is a light-weight model that performs most of the convolution operation in the perspective view, and the training and inference are significantly lighter and faster than all methods mentioned above.
Moreover, the training process of YOLOStereo3D does not depend on point-cloud data.

\subsubsection{Monocular 3D Object Detection}

Monocular 3D object detection is an ill-posed problem, but it provides many insights into how depth information can be estimated from a single image.
Tom \textit{et al.} \cite{tom2019howdepth} demonstrated that a typical monocular depth estimation network mainly estimates depth from the vertical position of an object.
The authors \cite{tom2019howdepth} provided the theoretical background for pseudo-LiDAR in monocular detection \cite{Vianney2019RefinedMPL}\cite{Weng2019Plidar}.
YOLOStereo3D is built upon the inference structure of M3D-RPN \cite{Brazil2019M3DRPN} and GAC \cite{liu2021GAC} and further enhances the final features with stereo matching results.

\section{Methods}
\begin{figure*}
    \centering
    \includegraphics[{width=1.0\textwidth}]{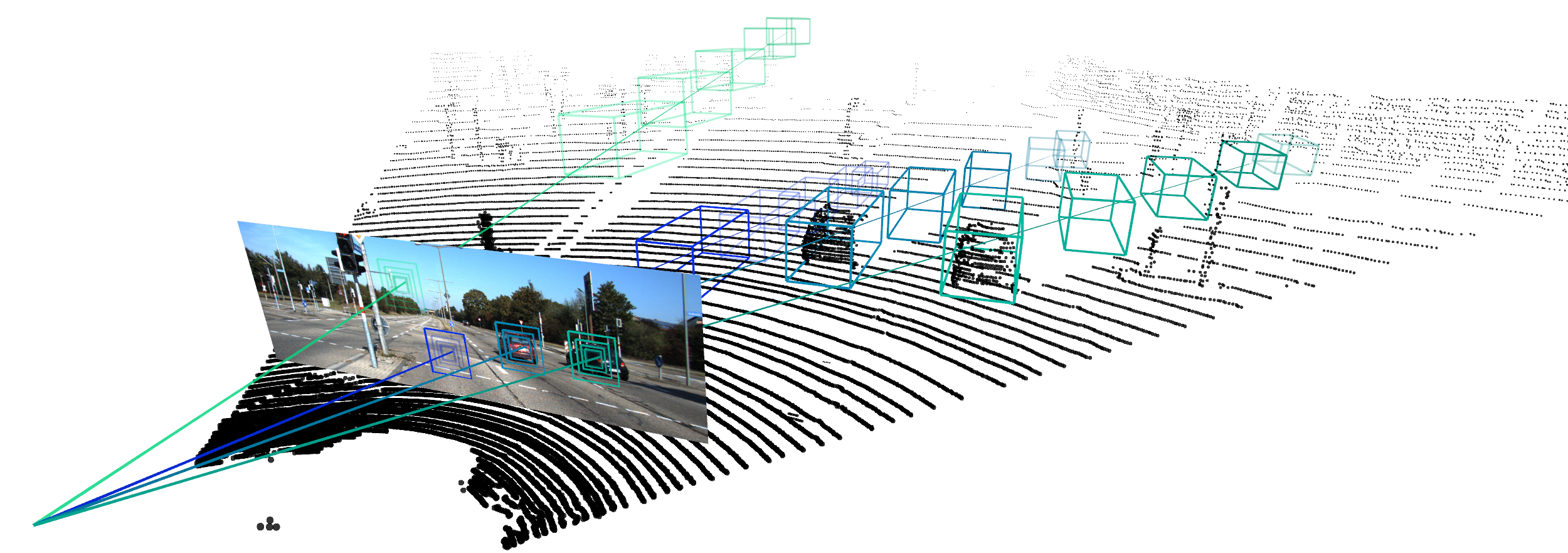}
        
    \caption{We project the center of each anchor box from the left image plane to 3D with its mean distance $\hat z$. We visualize the projected 3D bounding boxes with the mean width/height/length of the cars.
    We filter out anchors that are far from the ground plane during training (transparentized in the figure).
    Point clouds are displayed to indicate 3D positions in the figure. Best viewed in color.
    }
    \label{fig:anchor}
\end{figure*}

In this section, we elaborate on the network structure and methods applied in this paper.
First, we introduce the output definition and data-preprocessing tricks imported into and optimized for YOLOStereo3D \cite{liu2021GAC}.
Second, we re-introduce the light-weight cost volume that speeds up stereo matching and present the hierarchical densely-connected structure that fully exploits such thin features.
Finally, we deliver the loss function as well as the training and inference scheme of YOLOStereo3D.

\subsection{Anchors Definition and Preprocessings}
\label{sec.mono_anchor}
Since we adopt the inference structure of a monocular 3D object detection framework, we need to import the basic definition of anchors and we propose multiple optimized processing methods.
In this subsection, we present some of the preprocessing on the input and output of the network.
\subsubsection{3D Anchors and Statistical Priors}

Each anchor is described by 12 regressed parameters including
$[x_{2d}, y_{2d}, w_{2d}, h_{2d}]$ for the 2D bounding boxes;
$[c_x, c_y, z]$ for the 3D centers of objects on the left image;
$[w_{3d}, h_{3d}, l_{3d}]$ corresponding to the width, height and length of the 3D bounding boxes respectively;
and $[sin(2\alpha), cos(2  \alpha)]$ to estimate the observation angle/orientation of objects. 

We observe that $[sin(\alpha), cos(\alpha)]$  and $[sin(\alpha + \pi), cos(\alpha + \pi)]$ correspond to the same rectangular bounding box results in 3D object detection. As a result, we instead predict $[sin(2\alpha), cos(2\alpha)]$ in the regression branch.
We also add a classification channel to predict if $ |\alpha| > \frac{\pi}{2}$ to eliminate ambiguity, which intuitively means whether or not the object is facing the camera.

We incorporate 3D statistic priors into 2D anchors to improve the regression results.
To collect prior statistics of the anchors, we iterate through the training set, and for each anchor box of different shapes, collect all the objects assigned to this anchor based on the IoU metric.
Then, we compute the mean and the variance of $z$, $sin(2\alpha)$, $cos(2\alpha)$ for each box. 

Furthermore, we explicitly exploit scene-specific knowledge for autonomous vehicles and utilize the statistical information from the anchor boxes.
During training, we project dense anchor boxes into 3D with the mean depth value $\hat z$ and filter out anchor boxes that are far away from the ground plane based on their, as displayed in Figure~\ref{fig:anchor}. 

 For multi-class training, since the statistics for different types of obstacles, e.g., cars and pedestrians, are significantly different, we compute 3D priors for each category, separately.
 During training, we filter out anchor boxes dynamically based on the categories assigned. During inference, we also filter out anchor boxes dynamically based on the anchors' local categorical predictions.

\subsubsection{Data Augmentation for Stereo 3D Detection}

Data augmentation is useful to improve the generalization ability in deep learning applications.
However, the nature of stereo 3D detection limits the number of possible augmentation choices.
We follow \cite{Brazil2019M3DRPN} to apply photometric distortion concurrently on binocular images.
We also follow \cite{Li2019Stereo} to apply random flipping online during training. 
Random flipping includes flipping both RGB images, flipping the position and orientation of objects, and then switching left/right images.

\subsection{Multi-Scale Stereo Feature Extraction}
\label{subsec: multiscale_fusion}
The extraction of stereo features is one of the most time-consuming parts for many pre-existing stereo 3D object detection algorithms.
In this subsection, we re-introduce the cost volume formulation based on dot-product/cosine-similarity and present the hierarchical structure to utilize these features effectively.

\subsubsection{Light-weight Cost Volume}
\label{sec.costvolume}
Current state-of-the-art stereo matching algorithms usually construct 3D cost volume with concatenation, where the module iteratively shifts the right feature map horizontally over the left feature map, and at each step, concatenate the two features at each overlapping pixel. For binocular feature maps with the shape $[B, C, H, W]$, the shape of the output tensor $f_i$ is $[B, 2\cdot C, max\_disp, H, W]$. In this paper, we follow \cite{Luo2016EffiStereo} and \cite{wang2020FADNet} to apply a normalized dot-product to construct a thin cost volume. Such a module compute \textbf{correlation} between two overlapping pixels of the feature maps instead. The shape of the output tensor $f_i'$ becomes $[B, max\_disp, H, W]$.

The stereo matching process can be much faster. Consider two input feature maps of $[1, 64, 72, 320]$, which is a common shape of a KITTI image down scaled by 4. The forward pass of concatenation-based cost volume construction takes about 200 ms while the correlation-based cost volume takes about 7 ms on an Nvidia-1080Ti.

However, the number of output channels is smaller, which could cause the network to be numerically skewed towards monocular features during the fusion stage and downsampling the stereo matching results could induce further information loss.
We ease these two problems with densely connected ghost modules \cite{Han2020Ghost} and a hierarchical fusion structure.

\subsubsection{Densely Connected Ghost Module}
\label{sec.ghost}
As mentioned in Section~\ref{sec.costvolume}, we need to expand the width of the features to guide the network to skewed towards features produced by stereo matching.

Han \textit{et al.} propose the ghost module, which is an efficient module to produce redundant features \cite{Han2020Ghost}. It applies depthwise convolution to produce extra features, which requires significantly fewer parameters and FLOPs.
We go one step further and densely concatenate the original input features with the output of the original ghost module, thereby tripling the number of channels. As indicated in Figure~\ref{fig:architecture}, the mauve blocks in (b) are the results from ghost module and others denote densly connected residuals. 

Such a module preserves more information before downsampling and also rebalances the number of channels between stereo features and monocular semantic features during the fusion phase. 

\subsubsection{Hierachical Multi-scale Fusion Structure}
\label{sec.hierachical}
To minimize the information loss during the stereo matching phase while keeping the computational time tractable, we engineer a hierarchical fusion scheme.
At the downsampling level of $\frac{1}{4}$ and $\frac{1}{8}$, we construct a light-weight cost volume of a max-disparity of 96 and 192, respectively.
As shown in Figure~\ref{fig:architecture}, they are fed into a densely connected ghost module, downsampled, and concatenated with features at a smaller scale.
At a downsampling level of $\frac{1}{16}$, we first downsample the number of channels with $1\times 1$ convolution.
We then construct a small concatenation-based cost volume (also flattened to be a 2D feature map) to preserve more semantic information from the right images.

This arrangement can also be justified with high-level reasoning.
Features with higher resolution are usually local features with higher frequency portions, which are suitable for dense and accurate disparity estimation. In contrast, features with low resolution contain semantic information at a larger scale. 

\subsection{Training Scheme and Loss Function}
\label{sec.training}
The overall network structure is presented in Figure~\ref{fig:architecture}.
% Binocular images are collated as batches and sent to the backbone network.
Multi-scale features from binocular images are extracted and fused into stereo features to construct hierarchical cost volumes.
The stereo feature map is concatenated with the last feature map of the left image and fed to the regression/classification branch.
The stereo feature map is also fed into a decoder to predict a disparity map trained with an auxiliary loss. The auxiliary loss can regularize the training process.

\subsubsection{Auxiliary Disparity Supervision in Training}

As pointed out by Chen \textit{et al.} \cite{Chen2020DSGN}, disparity supervision is important to improve detection performance.
We also observe a similar phenomenon in our framework.
Without disparity supervision, the network may not be guided to produce local features useful in stereo matching to fully utilize the geometric potential of binocular images, and the network could be trapped in a local minimum similar to that of a monocular detection network.

We upsample the output of the final stereo features to $[W/4, H/4]$, and supervise the prediction with a sparse "ground truth" disparity derived from the traditional block matching algorithm in OpenCV \cite{opencv_library} during training.
During evaluation and testing, this disparity estimation branch is disabled to improve efficiency.

Though the disparity from the block matching algorithm is coarse and sparse, we empirically show that it significantly improves the network's performance.
\begin{table}
    \centering
    \caption{3D object detection results on the KITTI test set on \textbf{Car}.   "*" indicates usage of point cloud data or pretrained disparity estimation module.}
    \begin{tabular*}{0.46\textwidth}{ |l|c|c|r|}
        \hline
        {\bf Methods} & {\bf Easy/Moderate/Hard}  & {\bf Time}  \\ \hline
        RT3DStereo\cite{Hendrik2019RT3DStereo}       &29.90 \%/23.28 \%/ 18.96 \%& 0.08s\\
        StereoRCNN\cite{Li2019Stereo}                &47.58 \%/30.23 \%/ 23.72 \%&0.30s\\
        Pseudo-LiDAR*\cite{Weng2019Plidar}           &54.53 \%/34.05 \%/ 28.25 \%&0.40s\\
        OC Stereo*\cite{Pon2019OCStereo}             &55.15 \%/37.60 \%/ 30.25 \%&  0.35s\\
        ZoomNet*\cite{xu2020Zoomnet}                 &55.98 \%/38.64 \%/ 30.97 \%&  0.35s\\
        Disp R-CNN(velo)*\cite{Sun2020DispRCNN}      &59.58 \%/39.34 \%/ 31.99 \%&  0.42s\\
        Pseudo-LiDAR++*\cite{You2019PLPP}            &61.11 \%/42.43 \%/ 36.99 \%&  0.40s\\
        DSGN*\cite{Chen2020DSGN}                     &73.50 \%/52.18 \%/ 45.14 \%&  0.67s\\
        \hline 
        \textbf{Ours YOLOStereo3D}  & \textbf{65.68 \%}/\textbf{41.25 \%}/ \textbf{30.42 \%}& \textbf{0.08s}\\
        \hline
    \end{tabular*}
    
    \label{tab:test_results}
\end{table}

\begin{table}
    \centering
    \caption{3D object detection results on the KITTI test set on \textbf{Pedestrians}.}
    \begin{tabular*}{0.462\textwidth}{ |l|c|r|}
        \hline
        {\bf Methods} & {\bf Easy/Moderate/Hard} & {\bf Time}  \\ \hline
        RT3DStereo\cite{Hendrik2019RT3DStereo}       &  3.28 \%/  2.45 \%/  2.35 \%& 0.08s\\
        OC Stereo*\cite{Pon2019OCStereo}             & 24.48 \%/ 17.58 \%/ 15.60 \%& 0.35s\\
        DSGN*\cite{Chen2020DSGN}                     & 20.53 \%/ 15.55 \%/ 14.15 \%& 0.67s\\
        \hline 
        \textbf{Ours YOLOStereo3D}  &  \textbf{28.49 \%}/ \textbf{19.75 \%}/ \textbf{16.48 \%}& \textbf{0.08s}\\
        \hline
    \end{tabular*}
    
    \label{tab:test_ped_results}
\end{table}

\subsubsection{Loss Function}

We apply focal loss \cite{Yun2018Focal}\cite{Lin2018Focal} on classification, and smoothed-L1 loss \cite{Girshick2015Fastrcnn} on bounding box regression.  We follow the scheme of \cite{Zhang2019AcfNet} to apply stereo focal loss on the auxiliary disparity estimation.
First, we compute the expected distribution of disparity with a hard-coded variance $\sigma = 0.5$:
$$
    P(d) =\operatorname{softmax}\left(-\frac{\left|d-d^{g t}\right|}{\sigma}\right) =\frac{\exp \left(-c_{d}^{g t}\right)}{\sum_{d^{\prime}=0}^{D-1} \exp \left(-c_{d^{\prime}}^{g t}\right)}.
$$
Where $d$ represents the disparity and $c_d$ indicates the predicted confidence at disparity $d$. Then, following \cite{Zhang2019AcfNet}, stereo focal loss is defined as:
$$\mathcal{L}_{S F}=\frac{1}{|\mathcal{P}|} \sum_{p \in \mathcal{P}}\left(\sum_{d=0}^{D-1}\left(1-P_{p}(d)\right)^{-\alpha} \cdot\left(-P_{p}(d) \cdot \log \hat{P}_{p}(d)\right)\right)$$
where $D$ is the max-disparity, $\alpha$ is the focus weight, $\mathcal{P}$ presents the set of pixels involved, and $P_p(d)$, $\hat P_p(d)$ represents the expected and predicted distribution map of disaparity $d$.

The final loss function is simply the sum of the three losses.

\section{Experiments}
\label{section:Experiments}

We evaluate our method on the KITTI Object Detection Benchmark \cite{Geiger2012KITTI}.
The dataset consists of 7,481 training frames and 7,518 test frames.
Chen \textit{et al.} \cite{Chen2015kittisplit} further split the training set into 3,712 training frames and 3,769 validation frames. 
 In this section, we provide further training details and show the performance of YOLOStereo3D on the test set to compare it with existing models.
 \subsection{Implementation and Training Details}

 Modern deep learning frameworks are sensitive to hyperparameters choices, and critical design choices could profoundly influence the final performance.
 We introduce some crucial design choices before showing the performance, and the code will be made open source upon publication.

 We first determine the structure and the hyperparameters of the network on Chen's split \cite{Chen2015kittisplit}.
 Then, we retrain the final network on the entire training set with the same hyperparameters before uploading the results for testing onto the KITTI server. 
 An ablation study is also conducted on the validation set of Chen's split.
 
 The backbone of the network is ResNet-34 \cite{He2015Resnet}.
 The top 100 pixels of each image are cropped to speed up inference and training. The cropped input images are scaled to $288 \times 1280$.
 The network is trained with a batch size of 4 on a single Nvidia 1080Ti GPU (it takes about 7 GB of GPU memory, significantly less than other SOTA stereo detection algorithms) for 50 epochs on the KITTI training dataset.
 During inference, the network is fed one image at a time, and the total average processing time, including file IO, is about 0.08 s per frame. In contrast, most other stereo-based networks in the KITTI benchmark are several times slower.

 \subsection{Results on Test Set}

 The results are presented in Table~\ref{tab:test_results} alongside those of other state-of-the-art stereo 3D detection algorithms.

 The proposed YOLOStereo3D is fast and outperforms many pseudo-LiDAR methods or local point cloud methods and is the best performing algorithm without LiDAR usage.
 It also outperforms DSGN \cite{Chen2020DSGN} on pedestrian detection without an additional training schedule.

 \begin{table}
    \centering
    \caption{\textbf{Monocular} 3D object detection results of Cars on the KITTI \textbf{test} set.}
    \begin{tabular*}{0.47\textwidth}{  | l|c|c| }
        \hline
        {\bf Methods} & {\bf $\text{IoU}\ge 0.7$ 3D Easy/Moderate/Hard } & {\bf Time}\\ \hline
        M3D-RPN\cite{Brazil2019M3DRPN}      & 14.76 \% /       9.71 \%       / 7.42 \%  & 0.16s\\
        RTM3D\cite{Li2020RTM3DRM}           & 14.41 \% / 10.34 \% / 8.77 \%   & 0.05s\\
        AM3D\cite{Ma2019AM3D}               & 16.50 \% / 10.74 \% / 9.52 \%   & 0.40s\\
        D4LCN\cite{Ding2019D4LCN}           & 16.65 \% / 11.72 \% / \textbf{9.51} \% & 0.20s\\
        \hline
          \textbf{Ours}           & \textbf{19.24 \%}/  \textbf{12.37 \%}/   8.67 \%  & 0.05s\\
          \hline
  
    \end{tabular*} 
    \label{tab:test_mono}
  \end{table}
 
 \subsection{Test results for Monocular 3D Setting}
 To verify the effectiveness of the proposed anchor pre-processing techniques, we further test them in the task of monocular 3D object detection.
 Recall that we claim YOLOStereo3D being a monocular detector enhanced with stereo features. By taking away the image from the right camera, the multi-scale fusion module, and the disparity estimation branch, we obtain a standalone monocular detector. We enhance the backbone to be ResNet-101 \cite{He2015Resnet}. Following the proposed YOLOStereo3D, we compute the statistic for each anchor box and filter out deviated anchor boxes during training. 

 We also follow M3D-RPN \cite{Brazil2019M3DRPN} to post-process the prediction results to maximize the 2D-3D coherence. Notice that in YOLOStereo3D, we empirically find this post-processing step deteriorate the final performance, but it is beneficial in the monocular setting.
 
 The results are presented at Table~\ref{tab:test_mono}. 
 As shown in Table~\ref{tab:test_mono}, the proposed framework achieves state-of-the-art performance in KITTI Object Detection Benchmark under the monocular setting. The running time of the proposed monocular detector is about 50 ms per frame.

 \section{Model Analysis and Discussion}
 \label{section:Discussion}

 In this section, we further analyze the performance of YOLOStereo3D and discuss the effectiveness of several important design choices. 
 The baseline model here is only trained on the "Car" type.
We first conduct an ablation study to validate the contribution of anchor preprocessing, hierarchical fusion, and the densely connected ghost module on the validation set.
Then, we present and discuss some qualitative results.

\begin{table}
    \caption{Ablation study results of cars on the KITTI validation set}
    \begin{tabular*}{0.505\textwidth}{  |l|c| }
        \hline
        {\bf Methods} & {$\text{IoU}\ge 0.7$ 3D Easy/Moderate/Hard } \\ \hline
        % \textbf{YOLOStereo3D}           & \textbf{73.62 \%}/  \textbf{47.10 \%}/   \textbf{35.87 \%}  \\
        \textbf{YOLOStereo3D}               & \textbf{72.06 \%}/  \textbf{46.58 \%}/   \textbf{35.53 \%}  \\
        \hline
        w/o Anchor Prior        & 65.09 \%/   41.38 \%/  30.90 \% \\
        w/o Anchor Filtering    & 71.37 \%/   45.03 \%/  35.83 \% \\
        \hline
        w/o Channel Expand      & 64.16 \%/   39.96 \%/   30.02 \% \\
        w Naive Channel-expand  & 70.70 \%/   45.74 \%/   34.87 \% \\
        \hline
        w/o Scale 8             & 70.80 \%/   45.71 \%/   35.86 \% \\
        w/o Scale 16            & 68.64 \%/   44.54 \%/   33.95 \% \\
        \hline
        w/o Disparity supervision& 62.58 \%/   39.09 \%/   30.34 \% \\
        w PC supervision         & 72.05 \%/   46.59 \%/   35.62 \% \\
        \hline
        \textbf{YOLOMono3D}               & \textbf{21.66 \%}/  \textbf{14.20 \%}/   \textbf{11.07 \%}  \\
        \hline
        Mono w/o Prior        & 19.90 \%/   13.36 \%/   9.68 \% \\
        Mono w/o Filtering    & 20.50 \%/   13.45 \%/  10.50 \% \\
    \hline
    \end{tabular*} 
    \label{tab:ablation_study}
\end{table}

\begin{figure*}
    \centering
  \includegraphics[width=1.0\textwidth]{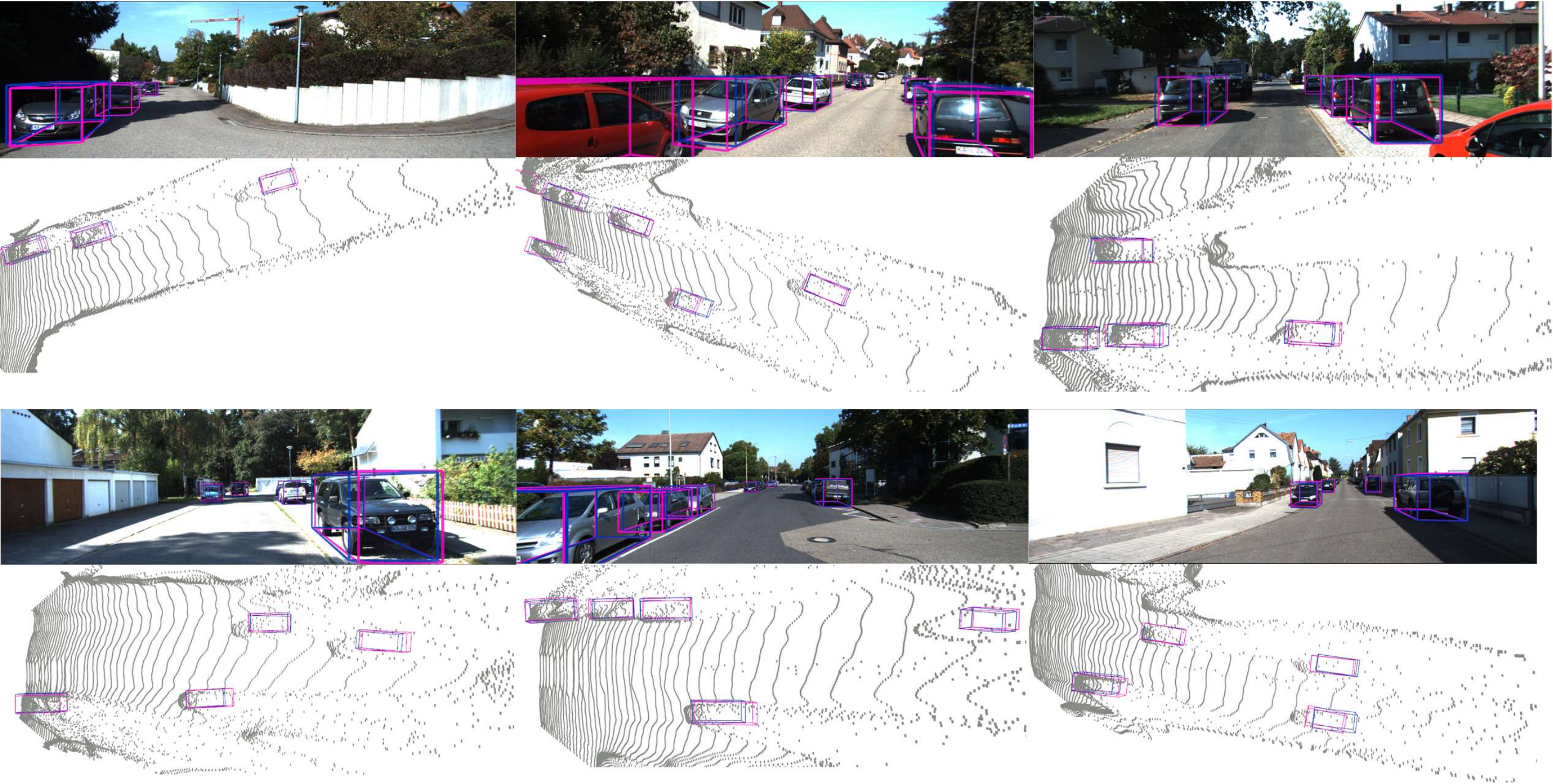}
    \caption{Qualitative examples from the validation set. The RGB images show the detection results and ground truth 3D bounding boxes on the left images. The bird's eye view images show the disparity prediction from the networks, along with detection results.
    The blue bounding boxes are 3D predictions from YOLOStereo3D, the pink bounding boxes are ground truth 3D bounding boxes, and point clouds are predictions from the disparity estimation branch of YOLOStereo3D.
    }
    \label{fig:examples}
\end{figure*}

\subsection{Ablation Study}

\subsubsection{Anchor Preprocessing}
We first test the effectiveness of including statistical information in each anchor.
In the first experiment, instead of predicting a depth value normalized by the depth prior, the network outputs a transformed depth output $\hat z$, where $z = 1/\sigma(\hat z) - 1$, following \cite{Chen2020MonoPair}.
In the second experiment, we do not filter out anchors during training, and the training loss is evaluated with all anchors. We conduct these two experiments in both the stereo setting and the monocular setting. The results are presented in Table~\ref{tab:ablation_study} respectively.

From the two table, we can observe that anchor priors significantly boost the performance of the network, and filtering out irrelevant anchors during training is also helpful. 
The performance gain can be observed in both monocular 3D detection and stereo 3D detection.
We suggest that we can ease the difficulty of depth inferencing by properly defining and preprocessing anchors specifically for 3D scene understanding in autonomous driving.

The improvement we apply on anchors can also be applied and verified in monocular 3D detection.
We further provide ablation experiments to validate the effectiveness of these processing methods under monocular 3D detection setting.

% \begin{table}
%     \centering
%     \caption{Ablation study results of cars on the KITTI validation set under the monocular setting.}
%     \begin{tabular*}{0.46\textwidth}{ | l|c |}
%         \hline
%         {\bf Methods} & {\bf $\text{IoU}\ge 0.7$ 3D Easy/Moderate/Hard } \\ \hline
%         \textbf{Ours}               & \textbf{21.66 \%}/  \textbf{14.20 \%}/   \textbf{11.07 \%}  \\
%         \hline
%         w/o Anchor Prior        & 19.90 \%/   13.36 \%/  9.68 \% \\
%         w/o Anchor Filtering    & 20.50 \%/   13.45 \%/  10.50 \% \\
%     \hline
%     \end{tabular*} 
%     \label{tab:ablation_study_mono}
%   \end{table}

\subsubsection{Densely Connected Ghost Module}
Densely-connected ghost modules are useful in expanding the number of channels in stereo processing.
We conduct two experiments to verify its effectiveness.
In the first experiment, we use a BasicBlock in resnet \cite{He2015Resnet} to replace the ghost module without expanding the number of channels, resulting in fewer channels during the fusion between RGB features and stereo features.
In the second experiment, we directly upsample the number of channels with $1\times 1$ convolution before feeding the tensor into a BasicBlock. 

We can observe from Table~\ref{tab:ablation_study} that the densely-connected ghost module is useful in improving the network's capability. From the first experiment, we demonstrate that expanding the number of channels is crucial for the network's performance.
In the second experiment, we further show that the densely-connected ghost module is better at preserving information than the naive $1\times 1$ convolution.

\subsubsection{Hierachical Fusion}
We respectively disable the stereo matching output on scale 8/16 to produce two networks to justify the usage of multi-scale fusion. We can observe from Table~\ref{tab:ablation_study} that the results of the two ablated models are inferior to that of the baseline model. 
We also point out that the forward pass of stereo matching modules on scale 8/16 is several times faster than that on scale 4.

As a result, we argue that hierarchically fusing stereo features from scale 8 and 16 is worth the effort. 
The baseline structure of hierarchical fusion in YOLOStereo3D achieves a fair balance between speed and performance.

\subsubsection{Disparity Supervision}
We also have an ablation study on the importance of disparity supervision. 
Similar to the conclusion in DSGN \cite{Chen2020DSGN}, disparity supervision significantly boosts the performance of the network. 

In the experiment, we show that such supervision is essential, but the results are not sensitive to the accuracy of the "target" disparity map.
The insight is that the network may only need slight regularizations in stereo matching submodules. The auxiliary loss is required to drive the network from falling back to a naive local optimal of monocular 3D object detection.

\subsection{Qualitative Results}
We show qualitative validation results in Figure~\ref{fig:examples}. 
The model displayed is YOLOStereo3D sharing the same hyperparameters as the model submitted to the KITTI server, but it is only trained on the training sub-split.

From the RGB images, we can observe that most of the successful predictions of YOLOStereo3D are visually consistent with the context. As shown in the bird's-eye-view images, though the disparity estimation may not correctly align with the ground truth 3D bounding boxes, the bounding box predictions from YOLOStereo3D are still reasonably accurate. 

The examples suggest that priors in anchor heads and the fusion between stereo matching features and RGB features could help the network to produce more visually consistent predictions and make the network more robust against potentially misleading disparity matching results.

\section{Conclusion}
\label{section:Conclusion}
In this paper, we presented YOLOStereo3D, an efficient stereo 3D object detection framework.
This work's major contribution is to take a step back to consider stereo 3D object detection as an enhanced monocular detection problem, rather than as an inaccurate LiDAR-based detection problem.
To achieve this, we first incorporated knowledge from real-time monocular 3D object detection frameworks and used priors in anchors for depth inference.
Then, we introduced the point-wise correlation module into the detection problems.
Finally, we used a hierarchical fusion framework that balances information preservation and computational burden.
We tested YOLOStereo3D on the KITTI Object Detection Benchmark.
The model produces competitive results among stereo frameworks while running at more than ten frames per second without any usage of LiDAR data.

It should be noted that we are converting the features from the right image to the left image.
In other words, the computational roles of the two images are not equal. Information loss in the right image is significant.
As a result, when an object is occluded in the left image but is more visible in the right image, the model could be significantly sub-optimal.

Nevertheless, since the model can achieve a competitive result with only one GPU and a short training time, YOLOStereo3D lowers the bar of stereo detection research. With a significantly faster inference speed and competitive performance, YOLOStereo3D can also boost the deployment of stereo setups on self-driving cars and mobile robots in the future.
%%%%%%%%%%%%%%%%%%%%%%%%%%%%%%%%%%%%%%%%%%%%%%%%%%%%%%%%%%%%%%%%%%%%%%%%%%%%%%%%

%%%%%%%%%%%%%%%%%%%%%%%%%%%%%%%%%%%%%%%%%%%%%%%%%%%%%%%%%%%%%%%%%%%%%%%%%%%%%%%%

% \section*{ACKNOWLEDGMENT}

% This work was supported by the National Natural Science Foundation of China, under grant No. U1713211, the
% Research Grant Council of Hong Kong SAR Government,
% China, under Project No. 11210017, No. 21202816, and the
% Shenzhen Science, Technology and Innovation Commission
% (SZSTI) under grant JCYJ20160428154842603, awarded to
% Prof. Ming Liu.
\addtolength{\textheight}{-7cm}
\bibliographystyle{IEEEtran}
\bibliography{reference}
\balance
%%%%%%%%%%%%%%%%%%%%%%%%%%%%%%%%%%%%%%%%%%%%%%%%%%%%%%%%%%%%%%%%%%%%%%%%%%%%%%%%

\end{document}